\pdfoutput=1 
\documentclass[11pt]{article}

\usepackage[font=libertinus, citestyle=numeric]{kurbanlab}
\usepackage{svg}
\usepackage{booktabs}

\usepackage{tabularx}
\usepackage{array}
\usepackage{times}
\usepackage{latexsym}
\usepackage[T1]{fontenc}
\usepackage[utf8]{inputenc}
\usepackage{microtype}
\usepackage{caption}
\usepackage{amsmath,amssymb,amsthm}
\usepackage{graphicx}
\usepackage{array}
\usepackage[table]{xcolor}
\usepackage{tikz}
\usepackage{siunitx}
\usepackage{multirow}
\usepackage{soul}
\usepackage{tabularx}
\usepackage{svg}
\usepackage{booktabs}
\usepackage{multirow}
\usepackage[flushleft]{threeparttable}
\definecolor{runone}{HTML}{EAF2FF}
\definecolor{runtwo}{HTML}{F1ECFF}
\definecolor{runthree}{HTML}{EAF8F0}
\usepackage{pifont}
\newcommand{\cmark}{\ding{51}}
\newcommand{\xmark}{\ding{55}}
\newcommand{\runone}{%
  \begingroup
  \setlength{\fboxsep}{2.5pt}%
  \colorbox{runone}{\strut\textbf{Run 1}}%
  \endgroup
}

\newcommand{\runtwo}{%
  \begingroup
  \setlength{\fboxsep}{2.5pt}%
  \colorbox{runtwo}{\strut\textbf{Run 2}}%
  \endgroup
}

\newcommand{\runthree}{%
  \begingroup
  \setlength{\fboxsep}{2.5pt}%
  \colorbox{runthree}{\strut\textbf{Run 3}}%
  \endgroup
}

\definecolor{PredGray}{gray}{0.94}
\definecolor{ObsBlue}{RGB}{232,241,250}
\definecolor{ObsBlueStrong}{RGB}{203,223,242}

\DeclareAffiliation{hbku}{%
  College of Science and Engineering, Hamad Bin Khalifa University, Doha, Qatar}

\DeclareAffiliation{tamuq}{%
  Department of Computer and Electrical Engineering,
  Texas A\&M University at Qatar, Doha, Qatar}

\DeclareAffiliation{tamu}{%
  Department of Computer and Electrical Engineering,
  Texas A\&M University, College Station, TX, USA}

\DeclareAffiliation{iub}{%
  Luddy School of Informatics, Computing, and Engineering,
  Indiana University Bloomington, Bloomington, IN, USA}

\newcommand{\dep}{\Delta}

\usetikzlibrary{arrows.meta,positioning,patterns}
\usepackage{threeparttable}
\title{Counterfactual Sensitivity Is Not Repairability: Auditing Replay Probes for Video Evidence} 
\Subtitle{CARVE: Counterfactual Auditing of Retrieved Visual Evidence} 
\RunningTitle{Counterfactual Sensitivity Is Not Repairability}

\Author{Rama AlHamidi}{tamuq}
\Author{Rasul Khanbayov}{hbku}
\Author{Erchin Serpedin}{tamu}
\Author[corresponding=hkurban@hbku.edu.qa, orcid=0000-0003-3142-2866]{Hasan Kurban}{hbku}


\CodeURL{https://github.com/KurbanIntelligenceLab/CARVE}

\begin{document}
\maketitle

\begin{abstract}
Tool-using video agents retrieve visual evidence before answering, but the
final answer is not forced to depend on what was retrieved. The natural black box test is counterfactual: destroy the semantic content of the frames the agent retrieved and check whether the answer changes, against a matched sham that re-executes the identical pipeline on those same frames. We introduce \textbf{CARVE}, a black-box counterfactual probe that compares answer changes under matched SHAM and DESTROY replays. Across three independent $k=3$ runs on a frozen VideoExplorer-style agent, DESTROY changes the answer 29.3 percentage points more often than SHAM, yielding a large and reproducible aggregate effect.
Question-level scores are less stable, and increasing the replay budget from
$k=3$ to $k=10$ reduces ties but weakens the original zero-threshold routing
policy. At $k=3$, CARVE selects 538 of 1,258 LVBench questions and improves
accuracy by 3.26 points, with higher fallback yield than most matched random
subsets. The score shows only a weak association with annotated temporal
coverage, so CARVE is best understood as a routing signal rather than a direct
grounding classifier. Our implementation is available at
\url{https://github.com/KurbanIntelligenceLab/CARVE}.
\end{abstract}

\printkeywords


\begin{figure}[t]
  \centering
  \kilgraphics{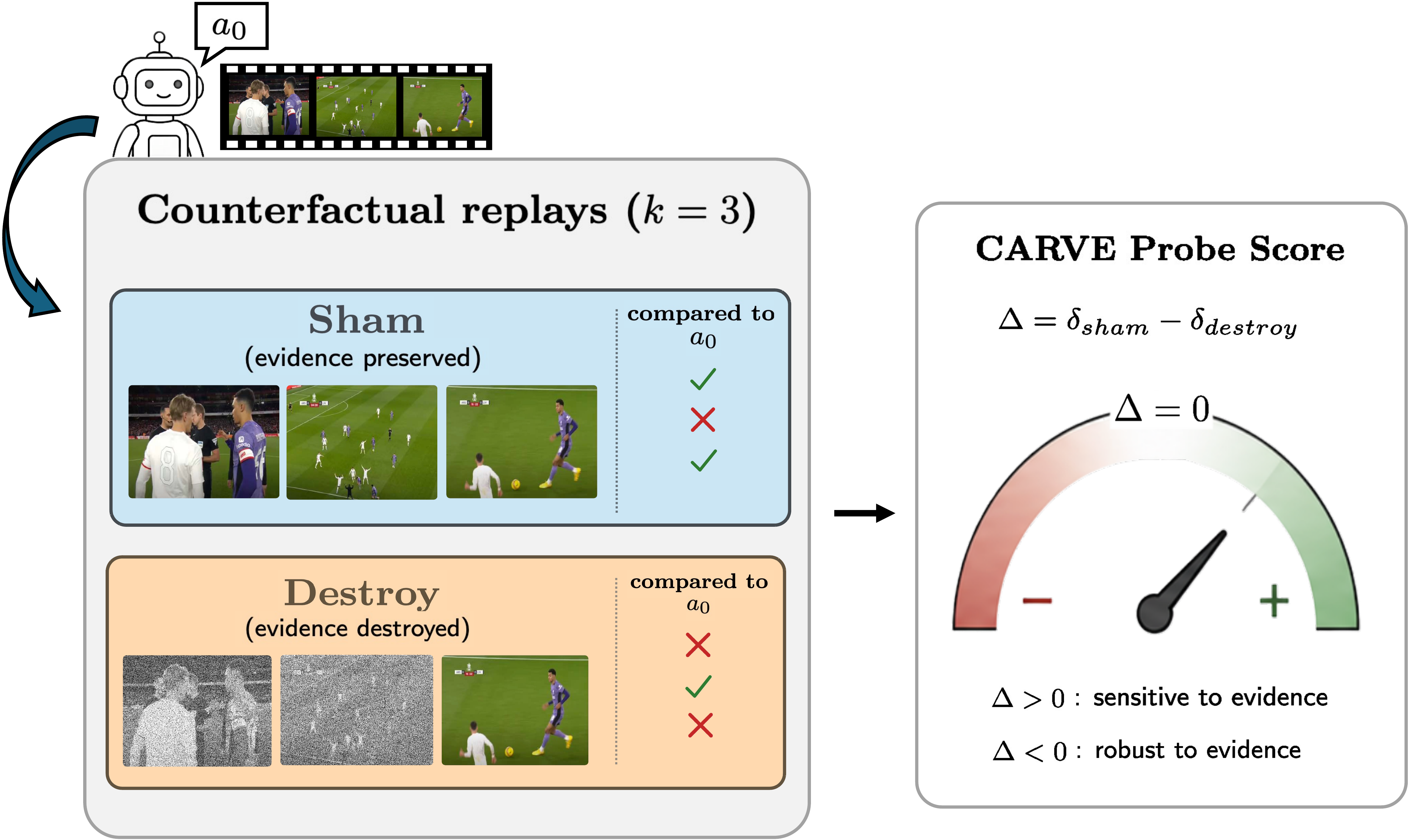}
  \caption{\textbf{CARVE estimates evidence dependence through matched counterfactual replays.} The agent produces a frozen answer $\hat a$ from retrieved evidence. CARVE replays the same question under two matched conditions: SHAM preserves the evidence, while DESTROY removes its visual content. Comparing replay answers with $\hat a$ gives change rates $\hat{\delta}_{\mathrm{sham}}$ and $\hat{\delta}_{\mathrm{destroy}}$. Their difference, $\hat{\Delta}$, measures answer change attributable to evidence destruction beyond SHAM variability: positive values favor DESTROY, zero indicates no separation, and negative values favor SHAM.}
  \label{fig:teaser}
\end{figure}

\section{Introduction}
\label{sec:intro}

A video agent answers by acting. It calls tools to ground a temporal window, retrieve frames, and render them as text, accumulating evidence before committing \citep{wang2024videoagent,wang2024videotree,yuan2025videoexplorer}. The architecture is meant to make the answer a reading of the gathered video, and nothing in it enforces that. Monolithic video reasoners are documented to answer from a language prior rather than from the frames \citep{luo2025drift}, and \citet{qiu2026videoseal} show that agentic systems produce correct answers unsupported by their own traces. Both diagnoses need annotated intervals or a judge model. We ask what a practitioner holding a fixed, possibly closed agent can measure with neither.

The natural test is interventional and runs on a frozen trajectory: take the frames the agent retrieved, destroy their semantic content, push them back through the unchanged pipeline, and count how often the committed answer changes. Counting flips alone proves nothing, because re-running a captioner on identical frames already changes answers. The comparison must be against a matched \emph{intact-frame sham}: same frames, same decoding, same prompt, same answerer, same number of replay draws, differing only in whether the frames keep their content. The excess flip rate under destruction over sham is the score $\dep$.

We built that probe and found a large, reproducible aggregate intervention
effect across independent replay runs. At the question level, however, the
score is only moderately stable, and small changes near zero can substantially
alter which questions are selected by a hard threshold. Under matched
independent replays, $\dep$ is estimated by a difference of two binomial
proportions, which determines its variance, discreteness, and finite-budget tie
probability. This analysis motivates a direct replay-budget test: increasing
$k$ resolves most low-budget ties and stabilizes the continuous score, but does
not necessarily improve the resulting routing policy. At $k=3$, the inclusive
nonpositive region nevertheless identifies a higher-yield fallback subset than
most matched random selections.

\paragraph{Contributions}
\begin{itemize}\itemsep1pt

\item \textbf{A matched counterfactual audit for frozen video agents.}
We introduce CARVE, a black-box probe that tests whether an agent's committed
answer depends on its retrieved visual evidence. CARVE contrasts semantic
evidence destruction with a matched intact-evidence sham, isolating
intervention sensitivity from the substantial variability caused by replaying
the pipeline itself. It requires no annotations, judge model, gradients,
logits, retraining, or modification of the underlying agent
(\S\ref{sec:probe}, \S\ref{sec:e0}).

\item \textbf{A finite-replay theory of what counterfactual probes can resolve.}
We show that the CARVE estimator is a difference of two binomial proportions,
which determines its sampling variance, discrete support, and probability of
exact ties. This analysis explains how a probe can be reliable in aggregate
while remaining unstable as a per-question decision rule, and provides a
general account of resolution limits in stochastic black-box intervention
tests (\Cref{thm:law}, \S\ref{sec:theory}).

\item \textbf{A direct separation of score resolution from decision utility.}
Using nested replay budgets, we show that additional replays resolve
low-budget ambiguity and stabilize the continuous score, yet can weaken a
fixed routing threshold. This establishes that more precise counterfactual
measurement does not necessarily produce a better controller, because policy
utility also depends on how actionable cases are distributed around the
decision boundary (\S\ref{sec:e1}).

\item \textbf{An empirical account of what makes the probe actionable.}
We disentangle temporal coverage, replay stability, destruction sensitivity,
and fallback repairability. CARVE is only weakly related to annotated temporal
coverage and should not be treated as a grounding certificate, but its
nonpositive region identifies a higher-yield fallback subset than matched
random routing. Secondary experiments further test selective prediction and
transfer across answerer backbones
(\S\ref{sec:e2}--\S\ref{sec:e4}).

\end{itemize}

\begin{kilkey} A large and reproducible counterfactual effect does not imply a stable question-level decision rule: increasing replay depth makes CARVE's score better resolved while weakening its original zero-threshold routing policy. \end{kilkey}

\section{Related work}
\label{sec:related}

\paragraph{Ungrounded video answers}
\citet{luo2025drift} show that chain-of-thought can lower video accuracy when
the reasoning trace drifts from the visual evidence, and introduce a Visual
Evidence Reward for post-training under GRPO. \citet{xu2026defacto} train for
evidence-grounded reasoning using positive, counterfactual, and random-masking
supervision in single-image settings. Both approaches modify model weights,
whereas CARVE audits a frozen system at inference time.

\paragraph{Diagnosing the agentic case}
\citet{qiu2026videoseal} is the closest prior work on evidence misalignment in
agentic long-video QA. They isolate the correct-but-ungrounded regime using
temporal overlap with annotated intervals and an LLM-judge audit, then separate
planning from answer authority and retrain the planner. Our temporal-coverage
analysis in \S\ref{sec:e2} follows their diagnostic framing. Their diagnostics
require gold intervals or a judge and their remedy modifies the agent, whereas
CARVE computes its score without either and wraps an already frozen system.

\paragraph{Perturbation as measurement}
Counterfactual interventions on chain-of-thought show that final answers can
survive corrupted reasoning
\citep{lanham2023measuring,rfeval2026}, while counterfactual input tests have
been used to evaluate explanation faithfulness
\citep{atanasova2023faithfulness}. In video, MECD masks events to recover causal
structure \citep{mecd2024}, while STEAR constructs patch-level temporal
counterfactuals within a Video-LLM \citep{fan2026stear}. Outside video, AgentSentry uses controlled counterfactual re-execution at
tool-return boundaries to diagnose causal takeover in tool-using agents
\citep{agentsentry2026}, and FACTOR perturbs non-causal visual attributes at
test time \citep{factor2026}. CARVE instead studies a stochastic, tool-using
agent through matched replay and characterizes the finite-budget variance,
discreteness, and threshold instability of the resulting per-question score.
\Cref{tab:novelty} in \Cref{app:position} provides a broader
comparison.

\paragraph{Selection, agreement, and abstention}
Self-consistency selects the modal answer across repeated samples
\citep{wang2023selfconsistency}, while soft variants relax exact-match voting
\citep{wang2024softsc}. ASC-MQRA combines voting with targeted re-watching of low-margin cases
\citep{kim2026ascmqra}. Reliable VQA frames uncertainty
through abstention and risk--coverage analysis
\citep{whitehead2022reliable}, with related mechanisms extended to video
\citep{yoon2025answerability}. Agreement and confidence measure output
stability or sharpness, whereas CARVE measures how the answer changes when the
retrieved visual evidence is destroyed relative to a matched sham. We compare
these signals directly in the routing and selective-prediction experiments.


\section{Setup and the matched counterfactual probe}
\label{sec:probe}

A video agent maps a question $q$ and video $v$ to a trajectory
\[
\tau=(u_1,o_1,\ldots,u_T,o_T)
\]
of tool calls and observations, together with a final answer $\hat a$.
We write $E\subseteq v$ for the \emph{retrieved evidence}, consisting of the
frames, temporal windows, and transcribed spans selected by the trajectory and
supplied to the final answering interface. CARVE treats the agent as a black
box exposing only $\tau$, $E$, and $\hat a$. It does not require access to
model weights, hidden states, or logits. The gold answer $a^\star$ is used only
for evaluation.

The agent studied here contains an intermediate visual-evidence renderer.
For replay condition $c$ and draw $j$, the answering interface is written as
\begin{equation}
\label{eq:interface}
z_c^{(j)}
=
h_{\omega_{c,j}}
\!\left(q,T_c(E)\right),
\qquad
\hat a_c^{(j)}
=
g\!\left(q,z_c^{(j)}\right),
\end{equation}
where $h$ converts the retrieved visual evidence into the representation
consumed by the final answerer, $g$ produces the multiple-choice answer, and
$\omega_{c,j}$ denotes residual replay variability. In our implementation,
this variability is dominated by regeneration of the rendered evidence.
The original answer $\hat a$ is frozen before probing and serves as the common
reference for every replay.

\begin{definition}[Matched SHAM and DESTROY interventions]
\label{def:mask}
The SHAM intervention preserves the retrieved visual content,
\[
T_{\mathrm{sham}}(E)=E,
\]
and re-executes the same preprocessing, rendering, parsing, and answering
pipeline.

The DESTROY intervention applies Fourier phase randomization to the same
retrieved visual evidence,
\[
T_{\mathrm{destroy}}(E)=m_{\mathrm{phase}}(E).
\]
The intervention preserves the evidence layout and amplitude spectrum while
disrupting phase-dependent visual structure.
\end{definition}

DESTROY is interpreted only relative to SHAM, which captures the variability
introduced by replaying intact evidence through the same pipeline.

\begin{assumption}[Matched independent replays]
\label{ass:matched}
For a fixed question, SHAM and DESTROY share the original question, retrieved
evidence locations, frame count, timestamps, resolution, ordering, prompts,
decoding settings, parsing procedure, and number of replay draws. The
conditions differ only in the visual content supplied to the renderer.
Replay draws are independent within each condition, and the two conditions are
executed using independent replay randomness.
\end{assumption}

For condition $c\in\{s,d\}$, define the population answer-change probability
\begin{equation}
\label{eq:flip_probability}
\delta_c(q)
=
\Pr_{\omega\sim\Omega_c}
\left[
\hat a_c^{(\omega)}\neq\hat a
\right],
\end{equation}
where $s$ denotes SHAM and $d$ denotes DESTROY.

\begin{definition}[Counterfactual score]
\label{def:drift}
The population counterfactual score is
\begin{equation}
\label{eq:population_delta}
\Delta(q)
=
\delta_d(q)-\delta_s(q).
\end{equation}
With $k$ replay draws per condition, it is estimated by
\begin{equation}
\label{eq:estimated_delta}
\hat{\Delta}(q)
=
\hat{\delta}_d(q)-\hat{\delta}_s(q),
\end{equation}
where
\begin{equation}
\label{eq:estimated_flip}
\hat{\delta}_c(q)
=
\frac{1}{k}
\sum_{j=1}^{k}
\mathbb{1}
\left[
\hat a_c^{(j)}\neq\hat a
\right].
\end{equation}
\end{definition}

Positive, zero, and negative values indicate whether DESTROY changes the answer
more often than, equally often as, or less often than SHAM. The score is not a
direct correctness or grounding label.

Because the two conditions use the same replay pipeline and the same number of
draws, their difference controls in expectation for variability arising from
rendering, answer regeneration, and pipeline re-execution. We additionally record an answer-self diagnostic that reruns only the final
answerer on fixed rendered evidence; it is used for analysis but is not part of
the CARVE score.

The primary operating point uses $k=3$ independent draws per condition. The
same definition is also evaluated at $k=10$ in the replay-budget experiment
and at $k=1$ in the compute-reduced cross-backbone analysis.


\section{What the probe can resolve}
\label{sec:theory}

The population score in Eq.~\eqref{eq:population_delta} is defined through
replay probabilities, but the observed score is estimated from finitely many
binary answer-change outcomes. This section characterizes the resulting
sampling variability and explains why stable aggregate behavior can coexist
with unstable per-question threshold decisions. All proofs are given in
\Cref{app:proofs}.

For notational simplicity, we suppress the question index and write
$\delta_s$ and $\delta_d$ for the SHAM and DESTROY answer-change probabilities.
Let
\[
b_j(p)=\binom{k}{j}p^j(1-p)^{k-j}
\]
denote the binomial probability mass function.

\begin{lemma}[Exact finite-replay law]
\label{thm:law}
Under \Cref{ass:matched},
\[
k\hat{\delta}_s\sim\operatorname{Bin}(k,\delta_s),
\qquad
k\hat{\delta}_d\sim\operatorname{Bin}(k,\delta_d),
\]
with the two counts independent. Therefore,
\[
\hat{\Delta}
\in
\left\{
-1,-1+\frac{1}{k},\ldots,
-\frac{1}{k},0,
\frac{1}{k},\ldots,
1-\frac{1}{k},1
\right\},
\]
and for every integer $m\in\{-k,\ldots,k\}$,
\begin{equation}
\label{eq:law}
\Pr\left[
\hat{\Delta}=\frac{m}{k}
\right]
=
\sum_j
b_{j+m}(\delta_d)\,
b_j(\delta_s),
\end{equation}
where terms with indices outside $\{0,\ldots,k\}$ are defined to be zero.
Moreover,
\begin{equation}
\label{eq:unbiased}
\mathbb{E}\left[\hat{\Delta}\right]
=
\delta_d-\delta_s
=
\Delta,
\end{equation}
and
\begin{equation}
\label{eq:variance}
\operatorname{Var}\left[\hat{\Delta}\right]
=
\frac{
\delta_d(1-\delta_d)+\delta_s(1-\delta_s)
}{k}.
\end{equation}
\end{lemma}

Equation~\eqref{eq:unbiased} shows that the estimator is unbiased for the
population destroy--sham difference. Equation~\eqref{eq:variance} shows that
its variance decreases at rate $1/k$, but can remain large for individual
questions when either replay probability is near one half. A reproducible
population mean therefore does not imply that each question receives nearly
the same estimate across independent runs.

\paragraph{Discreteness and exact ties}
The support of $\hat{\Delta}$ is spaced in increments of $1/k$. At the
primary budget $k=3$, the score can take only seven values, so exact empirical
ties occur with nonzero probability even when the population score is not
zero. Setting $m=0$ in Eq.~\eqref{eq:law} gives
\begin{equation}
\label{eq:tie_probability}
\Pr\left[\hat{\Delta}=0\right]
=
\sum_{j=0}^{k}
b_j(\delta_d)b_j(\delta_s).
\end{equation}
Substituting one population-wide pair of replay probabilities gives the
homogeneous tie prediction used in \Cref{sec:e1}. A discrepancy between
that prediction and the observed tie mass indicates that population-average
rates do not fully describe the question-level replay distribution.

\paragraph{Continuous agreement and threshold instability}
The routing policy converts the numerical score into
\[
R(q)=\mathbb{1}\!\left[\hat{\Delta}(q)\leq0\right].
\]
Thresholding discards information about score magnitude, so two runs can
produce moderately or strongly correlated continuous scores while disagreeing
substantially on which questions lie on either side of zero. Questions near the
threshold are especially sensitive to small replay-induced changes. Continuous
score agreement and routing-set agreement are therefore distinct empirical
properties.

\paragraph{Replay-budget implications}
Increasing $k$ reduces estimator variance and refines the score grid, so many
low-budget ties can move to either side of zero. This improves estimation of
the score itself, but does not guarantee that the same downstream threshold
becomes more useful. Routing utility also depends on how fallback benefit is
distributed around the decision boundary. More accurate estimation therefore
does not necessarily produce a better fixed routing policy;
\Cref{sec:e1} tests this distinction directly.

\section{Experimental design}
\label{sec:exp}

\paragraph{Agent and models}
We use the released VideoExplorer-style tool-using agent
\citep{yuan2025videoexplorer}, including its planner, temporal grounder, and
Qwen2.5-VL-7B-Instruct visual-evidence renderer
\citep{bai2025qwen25vl}. The planner selects tools and produces the final
multiple-choice answer. Complete software and run configurations are reported
in \Cref{app:repro}. Total GPU-hours could not be reconstructed
reliably because the experiments were distributed across several instances
and execution periods.

\paragraph{Benchmarks and evaluation universes}
LVBench \citep{wang2025lvbench} supplies temporal annotations and supports the
construct-validity analysis, while Video-MME \citep{fu2025videomme} provides a
second evaluation setting and the selective-prediction analysis.
The finalized LVBench universe draws on 90 of the 103 videos prepared for this study; questions from the remaining 13 videos failed to produce any scoreable output and are absent from the scored pool entirely. Starting from the resulting 1,308 scored questions (across those 90 videos), we exclude 42 questions without usable visual evidence and 8 without a valid parsed answer, leaving 1,258 eligible questions. The Video-MME universe contains 900 evaluated questions; 874 produce scoreable
vanilla answers. Among the full 900, 653 are eligible for the CARVE probe, 221 are ineligible but remain scoreable under the vanilla pipeline, 13 are
ineligible and unscored, and 13 terminate with an exception. The 653 eligible questions span 290 distinct videos. A question is probe eligible only when the original answer parses, the saved
evidence and timestamps can be reconstructed, the rendering prompt is
available, both counterfactual branches execute through the same replay
pipeline, and the evidence-bearing observations in the frozen answer context
can be intervened on. Gold answers and temporal annotations are used only for
evaluation and, where explicitly stated, development-set policy selection.

\paragraph{Counterfactual probe}
For each eligible question, we evaluate SHAM and DESTROY using $k=3$
independent replays per condition. SHAM preserves the retrieved visual content
while regenerating its pixel realization, whereas DESTROY applies phase
randomization to remove evidence-bearing visual structure. Each replay uses
the frozen question and retrieved evidence intervals from the original
trajectory. We retain the replay seed, rendered evidence, raw and parsed
answers, validity indicator, and answer-change indicator for every draw.
A compute-reduced $k=1$ variant is used only in the cross-backbone transfer
experiment. The replay-budget analysis extends the primary run from $k=3$ to
$k=10$ while preserving the original three draws and adding seven new draws
per condition.

\paragraph{Deferral policy and baselines}
The primary CARVE policy defers question $i$ to a secondary answerer when
$\hat{\Delta}_i\leq0$ and otherwise retains the vanilla answer. The
secondary answerer receives the frozen question and saved evidence context.
It does not begin a new agent trajectory, invoke retrieval or planning tools,
collect evidence outside the original evidence set, repeat the probe, or
abstain. We compare against Vanilla, Gated Look-more, final-answer self-consistency,
judge-grounding selection, and matched random deferral. Gated Look-more gives
selected questions an additional evidence-gathering rollout without using
$\hat{\Delta}$. Self-consistency replays the answering interface three
times on the frozen context and returns the modal answer. Judge-grounding
selection uses a text-only judge to prefer a Look-more candidate only when its
grounding score exceeds that of the vanilla candidate by a fixed margin.
Matched random deferral applies the same fallback operation to subsets with
the same cardinality as CARVE, with plain, video-balanced, and
evidence-matched sampling controls.

\paragraph{Inference cost}
All primary method comparisons use the same predefined inference budget.
Because CARVE divides this budget between counterfactual evidence rendering,
re-answering, and selective fallback, the methods need not make the same
number or type of model invocations. Each SHAM and DESTROY replay renders the
retrieved evidence items and then invokes the answerer. The finalized LVBench
population contains a mean of 2.693 visual-evidence items per question, so the
$k=3$ probe performs an average of 16.16 evidence-rendering invocations and
six answerer invocations per eligible question. These invocation counts
describe how CARVE allocates its budget and should not be interpreted as
equivalent units of computational cost. Exact end-to-end token totals for the probe are unavailable because the stored
SHAM and DESTROY records do not include per-invocation token telemetry.
Accordingly, the reported token counts cover the logged components of each
method, while the compute-matched comparison follows the fixed budget
allocation specified above.


\section{Results}
\subsection{The intervention is real in aggregate}\label{sec:e0}
We first assess probe stability across three independent runs, each using $k=3$ replays. \Cref{tab:seed_aggregate} summarizes the unchanged-evidence re-answering results across all three seeds. Across the three independent runs, DESTROY has a mean flip rate of 0.6389,
compared with 0.3455 for SHAM. Averaging the per-seed destroy--sham
differences gives a mean $\hat{\Delta}$ of 0.2934.

Thus, phase randomization changes the answer approximately 29.3 percentage
points more often, on average, than rerunning the same pipeline with intact
visual content.

\begin{table}[t]
  \centering
  \caption{\textbf{Aggregate probe behavior across three finalized replay seeds.} Entries are seed-level means and sample standard deviations across three independent $k=3$ runs on the same LVBench evaluation universe. Each seed is treated as an independent replication rather than pooled into a larger replay budget.}
  \label{tab:seed_aggregate}
  \small
  \begin{tabular}{@{}l
      S[table-format=1.4]
      S[table-format=1.4]@{}}
    \toprule
    \kilth{Metric} &
    {\kilth{Three-seed mean}} &
    {\kilth{Seed SD}} \\
    \midrule

    \multicolumn{3}{@{}l}{\textit{Probe flip rates}} \\
    SELF flip rate    & 0.1767 & 0.0068 \\
    SHAM flip rate    & 0.3455 & 0.0042 \\
    DESTROY flip rate & 0.6389 & 0.0027 \\

    \addlinespace
    \multicolumn{3}{@{}l}{\textit{Estimated-effect distribution}} \\
    Mean $\hat{\Delta}$      & 0.2934 & 0.0070 \\
    $\Pr(\hat{\Delta}<0)$    & 0.1258 & 0.0074 \\
    $\Pr(\hat{\Delta}=0)$    & 0.3086 & 0.0104 \\
    $\Pr(\hat{\Delta}>0)$    & 0.5656 & 0.0033 \\
    \bottomrule
  \end{tabular}
\end{table}

\subsection{Deeper replay resolves the score but weakens routing}
\label{sec:e1}

\begin{figure}
    \small
    \centering
    \kilgraphics{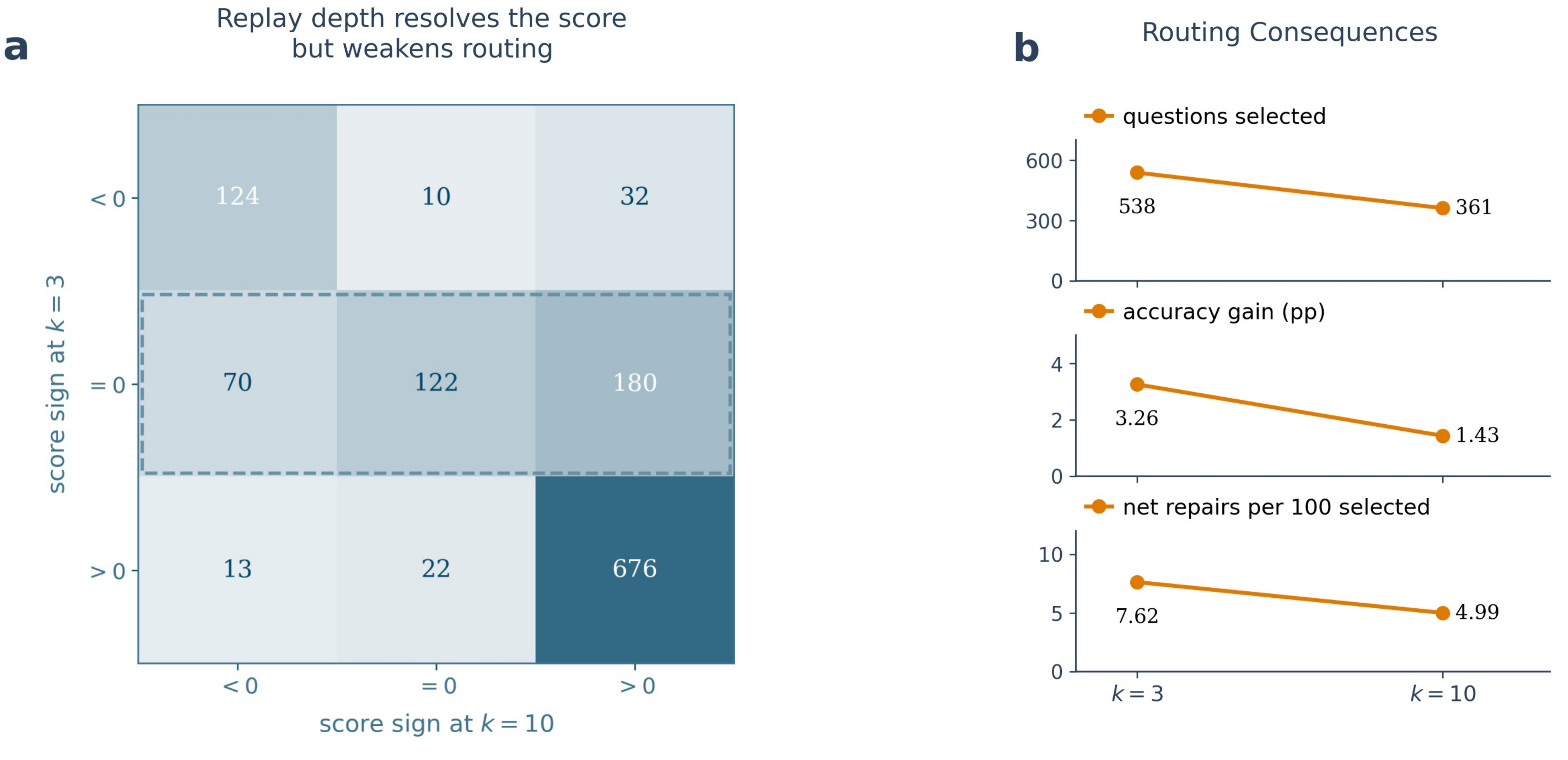}
    \caption{\textbf{Increasing replay depth resolves the score but weakens the fixed routing rule.}
    \textbf{(a)} Transition matrix for the 1,249 LVBench questions with valid
    scores at both $k=3$ and $k=10$. Most low-budget ties resolve, whereas only
    a small fraction of original non-ties become tied.
    \textbf{(b)} Consequences of applying the same
    $\hat{\Delta}\leq0$ routing rule at each replay budget. Increasing the
    budget selects fewer questions and reduces both final accuracy gain and
    net repairs per fallback call. Routing accuracy is evaluated over the full
    1,258-question population; questions without a valid probe score retain
    the vanilla answer.}
    \label{fig:replay-budget}
\end{figure}

\Cref{thm:law} gives the finite-budget distribution of
$\hat{\Delta}$ for specified SHAM and DESTROY flip probabilities. Using
the aggregate rates from \runone, it predicts the negative, zero, and positive
score mass expected from replay discreteness alone. \Cref{tab:tie_budget}
shows that the observed tie mass exceeds this homogeneous prediction at both
$k=3$ and $k=10$. Population-average flip rates therefore do not fully
describe the question-level score distribution. Increasing the replay budget substantially reduces this tie mass. Extending
\runone\ from $k=3$ to $k=10$ resolves most questions that were tied at the
lower budget while preserving the broad continuous score structure. However,
the additional resolution does not improve the original
$\hat{\Delta}\leq0$ routing rule. The transition pattern shows that additional replays primarily resolve
existing ties rather than creating new ones. Of the 372 questions tied at
$k=3$, 70 become negative, 122 remain tied, and 180 become positive at
$k=10$. In contrast, only 3.6\% of questions that were nonzero at $k=3$
become tied at $k=10$. The reduction in tie mass therefore reflects the
resolution of low-budget cases rather than a general contraction of scores
toward zero. The continuous scores remain broadly aligned across budgets, but the induced
routing decisions are less stable. The Jaccard similarity between the question
sets selected by $\hat{\Delta}\leq0$ is 0.569, showing that a substantial
fraction of questions changes operational status when the replay budget
increases. \Cref{fig:replay-budget} shows that the higher replay budget improves
score resolution but not the fixed zero-threshold policy. At $k=10$, the
inclusive rule selects fewer questions and produces a smaller accuracy gain
and lower repair yield. Its matched-random percentile also falls sharply,
indicating that the higher-budget selected set no longer separates clearly
from rate-matched random routing.

\begin{table}[t]
    \centering
    \caption{
    \textbf{The probe has a small aggregate association with temporal coverage, but weak question-level discrimination.} Statistics are computed over LVBench questions with valid probe scores; 95\% confidence intervals are obtained by bootstrap resampling at the video level. Gold coverage denotes positive temporal overlap between retrieved evidence and the annotated answer-bearing interval. Among correct answers, covered questions receive higher mean scores than uncovered questions, while the association reverses among incorrect answers. Correlations with temporal recall and IoU are small, and the ROC-AUC is close to chance.
    }
    \fontsize{8}{9.5}\selectfont
    \setlength{\tabcolsep}{2.8pt}
    \renewcommand{\arraystretch}{1.02}
    \begin{tabular*}{\textwidth}{@{\extracolsep{\fill}}llrcc@{}}
    \toprule
    \multicolumn{5}{@{}l}{\emph{Score by evidence coverage and correctness}} \\
    \kilth{Gold cov.} & \kilth{Correct} & \kilth{$n$} & \kilth{Mean $\hat{\Delta}$} & \kilth{95\% CI} \\
    \midrule
    Yes & Yes & 275 & 0.453 & [0.403, 0.502] \\
    No  & Yes & 293 & 0.325 & [0.266, 0.387] \\
    Yes & No  & 245 & 0.191 & [0.136, 0.244] \\
    No  & No  & 436 & 0.242 & [0.207, 0.278] \\
    \midrule
    \multicolumn{5}{@{}l}{\emph{Association with evidence coverage}} \\
    \multicolumn{3}{@{}l}{Correct-answer mean difference}
        & +0.127 & [0.042, 0.209] \\
    \multicolumn{3}{@{}l}{Spearman, temporal recall}
        & +0.077 & [0.021, 0.131] \\
    \multicolumn{3}{@{}l}{Spearman, temporal IoU}
        & +0.087 & [0.031, 0.141] \\
    \multicolumn{3}{@{}l}{ROC-AUC}
        & 0.532 & [0.499, 0.566] \\
    \multicolumn{3}{@{}l}{PR-AUC}
        & 0.442 & [0.400, 0.497] \\
    \bottomrule
    \end{tabular*}
    \label{tab:e1}
\end{table}

Much of the $k=3$ policy's utility comes from the tied questions admitted by the inclusive threshold (\Cref{sec:e3}). With additional replays, many ties move away from zero; in particular,
180 become positive and are no longer selected. Thus, greater replay depth
produces a better-resolved score but a weaker decision rule. The $k=3$ policy
succeeds not because zero is a stable boundary between repairable and
non-repairable questions, but because its inclusive threshold happens to
retain a low-budget tie block with high empirical repair yield.

\subsection{The association with temporal coverage is weak and conditional}\label{sec:e2}
Among the 1,249 LVBench questions with valid scores, retrieved evidence
overlaps the gold interval for 520 questions, or 41.6\% of the evaluated
population. Moreover, 293 of the 568 correctly answered questions have no
temporal overlap with the annotated interval. Correctness and annotated
evidence coverage are therefore far from equivalent in this agent pipeline. \Cref{tab:e1} shows that the association between
$\hat{\Delta}$ and temporal coverage depends on whether the answer is
correct. Among correct answers, covered questions receive higher scores than
uncovered questions, with a video-clustered mean difference of $+0.127$. Among incorrect answers, however, the direction reverses, with covered questions receiving lower scores than uncovered
questions. Thus, the relationship between counterfactual sensitivity and
temporal coverage is conditional on answer correctness rather than uniform
across the evaluated population. The continuous overlap measures support the same limited conclusion. Correlations with temporal recall and IoU are positive but small, while the
ROC-AUC for distinguishing covered from uncovered questions remains close to chance and its confidence interval includes $0.5$. The PR-AUC is also only
slightly above the coverage prevalence of 0.416. Accordingly, $\hat{\Delta}$ exhibits a weak aggregate relationship with annotated coverage, but does not reliably determine whether an individual question's retrieved evidence overlaps the answer-bearing interval. More broadly, the score reflects a mixture of replay stability under SHAM and sensitivity to evidence destruction, rather than a direct measure of temporal grounding.

\paragraph{Component decomposition}
The aggregate score difference masks distinct behavior in its two
components (\Cref{tab:component-decomposition}). Among correct answers, covered questions have substantially lower SHAM flip
rates than uncovered questions, while their DESTROY rates differ much less.
Their higher $\hat{\Delta}$ therefore arises mainly from greater stability
under intact replay. Among incorrect answers, SHAM rates are nearly unchanged
by coverage, whereas covered questions have lower DESTROY flip rates. This
produces the reversed coverage association among wrong answers. The same
pattern persists at $k=10$, showing that it is not an artifact of low-budget
ties.

\begin{table}[t]
\centering
\caption{
\textbf{CARVE identifies a higher-yield fallback subset than matched random
routing.} Routing outcomes are computed over the 1,258-question LVBench evaluation universe. Repairs and harms are question counts; yield is net repairs per 100 fallback calls. Random rows report mean yield across matched random subsets containing the same number of questions as the CARVE-selected set.
}
\small
\setlength{\tabcolsep}{2pt}
\renewcommand{\arraystretch}{1.10}
\begin{tabularx}{\columnwidth}{
    @{}
    >{\raggedright\arraybackslash}X
    >{\centering\arraybackslash}p{0.14\columnwidth}
    >{\centering\arraybackslash}p{0.13\columnwidth}
    >{\centering\arraybackslash}p{0.12\columnwidth}
    >{\centering\arraybackslash}p{0.14\columnwidth}
    @{}
}
\toprule
\kilth{Policy} &
\kilth{Selected} &
\kilth{Repairs} &
\kilth{Harms} &
\kilth{\shortstack{Yield\\per 100}} \\
\midrule
$\hat{\Delta}<0$
& 166 & 33 & 19 & 8.43 \\

$\hat{\Delta}=0$
& 372 & 65 & 38 & 7.26 \\

$\hat{\Delta}\leq0$
& 538 & 98 & 57 & 7.62 \\
\midrule

Plain random
& 538 & -- & -- & 5.09 \\

Video-balanced random
& 538 & -- & -- & 5.28 \\

Evidence-matched random
& 538 & -- & -- & 5.30 \\
\bottomrule
\end{tabularx}
\label{tab:routing}
\end{table}

\begin{table}[t]
\centering
\caption{\textbf{Accuracy and rerouting outcomes.}
Results are evaluated at the question level on LVBench ($n=1{,}258$) and
Video-MME ($n=874$), with paired 95\% confidence intervals clustered by video.
$\Delta$Acc.\ is measured relative to Vanilla on the same evaluation universe.
For Random deferral, the starred interval is the central 95\% range of the
matched Monte Carlo random-selection distribution rather than a confidence
interval. Repairs and harms count question-level answer changes from incorrect
to correct and correct to incorrect, respectively. The probe issues no planner
tool calls, so equal tool counts do not imply equal total compute; token totals
also exclude some probe rendering.}
\small
\setlength{\tabcolsep}{5pt}
\begin{tabular}{
    @{}l
    S[table-format=2.2]
    S[table-format=1.2]
    r
    S[table-format=+1.2]
    S[table-format=3.0]
    S[table-format=3.0]
    l@{}
}
\toprule
\kilth{Method}
& \kilth{Acc.\ (\%)}
& \kilth{Tools}
& \kilth{Tok.}
& \kilth{$\Delta$Acc.}
& \kilth{Repairs}
& \kilth{Harms}
& \kilth{95\% CI} \\
\midrule

\multicolumn{8}{@{}l}{\textbf{LVBench} ($n=1{,}258$)} \\
Vanilla
& 45.47 & 4.21 & 32.2k
& {--} & {--} & {--} & {--} \\
Judge-grnd.
& 45.47 & 4.39 & 37.6k
& 0 & 3 & 3 & {\footnotesize[-0.40, +0.40]} \\
Look-more
& 45.07 & 4.39 & 37.5k
& -0.40 & 2 & 7 & {\footnotesize[-0.92, +0.07]} \\
Self-consist.
& 43.48 & 4.21 & 36.9k
& -1.99 & 32 & 57 & {\footnotesize[-3.58, -0.40]} \\
Random deferral
& 47.65 & 4.21 & 35.8k
& +2.18 & {--} & {--} & {[+0.95, +3.42]*} \\
CARVE
& 48.73 & 4.21 & 36.9k
& +3.26 & 98 & 57 & {[+1.23, +5.28]} \\
\midrule
\multicolumn{8}{@{}l}{\textbf{Video-MME} ($n=874$)} \\
Vanilla
& 47.83 & 5.43 & 50.7k
& {--} & {--} & {--} & {--} \\
Judge-grnd.
& 47.94 & 5.59 & 58.9k
& 0.11 & 2 & 1 & [-0.23, +0.57] \\
Look-more
& 48.17 & 5.59 & 58.9k
& +0.34 & 5 & 2 & [-0.23, +0.92] \\
Self-consist. 
& 48.52 & 5.43 & 55.9k
& +0.69 & 76 & 70 & [-1.95, +3.32] \\
CARVE
& 51.26 & 5.43 & 62.6k
& \multicolumn{1}{c}{\textbf{+3.43}}
& 74 & 44 & [+1.14, +5.72] \\
\bottomrule
\end{tabular}

\begin{tablenotes}[flushleft]
\footnotesize
\item
*For Random deferral, Acc. and $\Delta$Acc. are means over matched Monte Carlo selections; because no single realized routing set corresponds to these means, question-level repair and harm counts are omitted.
\end{tablenotes}

\label{tab:main}
\end{table}

\subsection{CARVE improves routing efficiency}
\label{sec:e3}
The primary $\hat{\Delta}\leq0$ policy selects 538 of 1,258 questions and
raises accuracy from 45.47\% to 48.73\%, a gain of 3.26 percentage points
(\Cref{tab:main}). Within this subset, fallback reconsideration
produces 98 repairs and 57 harms, corresponding to 41 net repairs, or 7.62
net repairs per 100 fallback calls. CARVE therefore obtains a meaningful
accuracy improvement while routing only 42.8\% of the evaluation population
to the fallback model.
\noindent
\textbf{Where does the gain come from?}
\Cref{tab:routing} decomposes the inclusive policy into its two
constituent strata. The strict $\hat{\Delta}<0$ region selects 166
questions and yields 14 net repairs, whereas the larger
$\hat{\Delta}=0$ stratum contributes a further 27. Thus, tied questions
account for 65.9\% of the policy's total improvement. The negative stratum has a slightly higher point-estimate yield than the tied
stratum, but the difference is modest and the negative subset is substantially
smaller. The results therefore do not establish that increasingly negative
scores imply greater repairability. Instead, both parts of the nonpositive
region contribute useful fallback cases: negative scores provide a somewhat
denser subset, while ties supply most of the total gain.

\paragraph{Comparison with matched random routing}
\Cref{tab:routing} compares CARVE with random subsets containing the same
number of questions. The comparison suggests that CARVE concentrates fallback benefit more
effectively than arbitrary routing, although the separation is not decisive:
its observed yield remains within each random control's 95\% range. We
therefore interpret the result as evidence of improved routing efficiency,
rather than as a definitive advantage over every matched-random policy.
Overall, the nonpositive-score region produces a higher-yield fallback subset
than the average matched control, while providing only limited evidence that
$\hat{\Delta}$ finely ranks questions by repairability.

\subsection{Selective prediction and cross-backbone transfer}
\label{sec:e4}

On the strict common-valid Video-MME subset, $\hat{\Delta}$ outperforms
sampled-position confidence in the high-coverage regime but not across the
full risk--coverage curve. In a separate cross-backbone experiment, the
inclusive routing rule produces positive mean transfer gains in both
directions, although the magnitude is asymmetric. These results suggest that
CARVE retains some utility beyond the primary LVBench experiment, but more
clearly as a coarse routing signal than as a universally superior ranking
score. Full results are reported in \Cref{app:secondary}.


\section{Conclusion}
CARVE audits whether a frozen video agent's answer responds to its retrieved
visual evidence by comparing matched SHAM and DESTROY replays. The
intervention produces a large and reproducible aggregate effect, while
finite-replay scores are less stable at the question level. Increasing the
replay budget resolves most low-budget ties but weakens the original
zero-threshold routing policy. CARVE should therefore be treated as a budget-dependent routing signal rather than a direct grounding certificate.


\section*{Limitations}
\paragraph{The score is not a correctness or grounding label}
CARVE measures the difference between answer changes under DESTROY and SHAM.
A wrong answer may depend strongly on retrieved evidence, while a correct
answer may arise from priors, alternate evidence, or chance. Likewise,
annotated temporal overlap is not equivalent to successful evidence use.
In our construct-validity analysis, the relationship between
$\hat{\Delta}$ and coverage depends on answer correctness and is driven by
different SHAM and DESTROY components across strata. The score therefore
should not be interpreted as a direct estimator of temporal grounding.

\paragraph{Finite replay limits per-question resolution}
At small $k$, $\hat{\Delta}$ lies on a coarse grid and has substantial
question-level variance. The aggregate destroy--sham effect is stable across
three independent runs, but continuous scores are only moderately reproducible
and hard trigger membership is less stable. Increasing the budget from
$k=3$ to $k=10$ reduces ties and improves continuous-score stability, but does
not improve the original zero-threshold routing rule. Replay-budget choice is
therefore part of the policy design rather than a purely statistical
precision decision.

\paragraph{The matched-replay model is an approximation}
\Cref{thm:law} assumes independent Bernoulli replay outcomes within and
across SHAM and DESTROY. Real model executions may exhibit dependence through
shared prompts, common evidence, deterministic preprocessing, or correlated
generation behavior. The exact finite-$k$ law should therefore be understood
as a conditional sampling model rather than a complete description of the
agent pipeline. The empirical excess tie mass at both $k=3$ and $k=10$ shows
that one population-wide pair of flip probabilities does not fully capture
question-level heterogeneity.

\paragraph{Phase randomization is an imperfect intervention}
DESTROY preserves frame count, timing, dimensions, ordering, container format,
and amplitude spectrum while disrupting phase-dependent visual structure.
It may nevertheless preserve residual cues or introduce artifacts that affect
the renderer independently of semantic evidence. We do not claim that
phase-randomized frames contain no usable information. The intervention is
interpreted only through comparison with the matched SHAM condition.

\paragraph{The intervention is applied to retrieved evidence only}
CARVE tests dependence on the evidence exposed in the frozen trajectory. It
does not perturb information retained in model parameters, textual priors,
earlier tool outputs outside the intervened evidence set, or alternative
visual evidence that the original trajectory did not retrieve. A low score
therefore does not prove that the answer is independent of the full video.

\paragraph{Gold temporal annotations are incomplete proxies for evidence use}
The construct-validity analysis treats overlap with the annotated
answer-bearing interval as temporal coverage. More than half of the correctly
answered LVBench questions lack such overlap, showing that correctness and
annotated coverage are not interchangeable in this pipeline. Some answers may
use alternate valid evidence, while some annotation intervals may not capture
all sufficient evidence. Consequently, the reported coverage associations
measure agreement with the benchmark annotation rather than direct access to
the model's causal reasoning process.

\paragraph{Cost telemetry is incomplete}
The primary comparisons follow the predefined inference-budget allocation,
but the methods distribute that budget across different invocation types.
CARVE uses evidence rendering, counterfactual re-answering, and selective
fallback, whereas other methods allocate the same budget differently.
The stored SHAM and DESTROY records do not contain per-invocation token counts,
so exact end-to-end CARVE token usage cannot be reconstructed. Reported token
totals cover the logged components and should not be interpreted as complete
probe cost.

\paragraph{The routing result is threshold specific}
The primary policy uses the inclusive rule
$\hat{\Delta}\leq0$. Its utility at $k=3$ depends substantially on the tied
stratum, which contributes most of the observed net gain. At $k=10$, many of
these cases move above zero and are excluded, reducing policy performance.
The results therefore validate this operating point rather than establishing
that zero is a universally optimal threshold or that more negative scores
monotonically imply greater repairability.

\paragraph{Generalization remains limited}
The main routing and construct-validity analyses use one VideoExplorer-style
agent and one primary answerer. Transfer is evaluated between two answerer
backbones within the same broader scaffold and at the compute-reduced
$k=1$ setting. The rule transfers with positive mean gain in both directions,
but the magnitude is asymmetric. Video-MME selective-prediction performance
also depends on the retained coverage range. Broader conclusions require
additional agents, rendering pipelines, intervention operators, and video
benchmarks.

\paragraph{Absolute performance is agent dependent}
This is primarily a within-agent study of counterfactual diagnosis and routing.
Absolute benchmark accuracy depends on the underlying retrieval, planning, and
answering components and may remain below stronger published systems. CARVE is
not intended to replace improvements to the base video agent.


\bibliography{references}

@article{luo2025drift,
      title={When Thinking Drifts: Evidential Grounding for Robust Video Reasoning}, 
      author={Luo, Mi and Xue, Zihui and Dimakis, Alex and Grauman, Kristen},
      year={2025},
      eprint={2510.06077},
      archivePrefix={arXiv},
      primaryClass={cs.CV},
      url={https://arxiv.org/abs/2510.06077}, 
}

@article{xu2026defacto,
  title={Defacto: Counterfactual thinking with images for enforcing evidence-grounded and faithful reasoning},
  author={Xu, Tianrun and Jing, Haoda and Li, Ye and Wei, Yuquan and Feng, Jun and Chen, Guanyu and Gao, Haichuan and Zhang, Tianren and Chen, Feng},
  journal={arXiv preprint arXiv:2509.20912},
  year={2025}
}

@article{fan2026stear,
  title={STEAR: Layer-Aware Spatiotemporal Evidence Intervention for Hallucination Mitigation in Video Large Language Models},
  author={Fan, Linfeng and Tian, Yuan and Li, Ziwei and Lu, Zhiwu},
  journal={arXiv preprint arXiv:2604.03045},
  year={2026}
}

@article{qiu2026videoseal,
  title={VideoSEAL: Mitigating Evidence Misalignment in Agentic Long Video Understanding by Decoupling Answer Authority},
  author={Qiu, Chenhao and Zhang, Yechao and Luo, Xin and Song, Shien and Liu, Xusheng},
  journal={arXiv preprint arXiv:2605.12571},
  year={2026}
}

@article{yuan2025videoexplorer,
  title={Videoexplorer: Think with videos for agentic long-video understanding},
  author={Yuan, Huaying and Liu, Zheng and Zhou, Junjie and Qian, Hongjin and Shu, Yan and Sebe, Nicu and Wen, Ji-Rong and Dou, Zhicheng},
  journal={arXiv preprint arXiv:2506.10821},
  year={2025}
}

@article{wang2023selfconsistency,
  title={Self-consistency improves chain of thought reasoning in language models},
  author={Wang, Xuezhi and Wei, Jason and Schuurmans, Dale and Le, Quoc and Chi, Ed and Narang, Sharan and Chowdhery, Aakanksha and Zhou, Denny},
  journal={arXiv preprint arXiv:2203.11171},
  year={2022}
}

@misc{bai2025qwen25vl,
      title={Qwen2.5-VL Technical Report}, 
      author={Shuai Bai and Keqin Chen and Xuejing Liu and Jialin Wang and Wenbin Ge and Sibo Song and Kai Dang and Peng Wang and Shijie Wang and Jun Tang and Humen Zhong and Yuanzhi Zhu and Mingkun Yang and Zhaohai Li and Jianqiang Wan and Pengfei Wang and Wei Ding and Zheren Fu and Yiheng Xu and Jiabo Ye and Xi Zhang and Tianbao Xie and Zesen Cheng and Hang Zhang and Zhibo Yang and Haiyang Xu and Junyang Lin},
      year={2025},
      eprint={2502.13923},
      archivePrefix={arXiv},
      primaryClass={cs.CV},
      url={https://arxiv.org/abs/2502.13923}, 
}

@inproceedings{wang2025lvbench,
  title={Lvbench: An extreme long video understanding benchmark},
  author={Wang, Weihan and He, Zehai and Hong, Wenyi and Cheng, Yean and Zhang, Xiaohan and Qi, Ji and Ding, Ming and Gu, Xiaotao and Huang, Shiyu and Xu, Bin and others},
  booktitle={2025 IEEE/CVF International Conference on Computer Vision (ICCV)},
  pages={22958--22967},
  year={2025},
  organization={IEEE}
}

@inproceedings{wang2024videoagent,
  title={Videoagent: Long-form video understanding with large language model as agent},
  author={Wang, Xiaohan and Zhang, Yuhui and Zohar, Orr and Yeung-Levy, Serena},
  booktitle={European Conference on Computer Vision},
  pages={58--76},
  year={2024},
  organization={Springer}
}

@inproceedings{wang2024videotree,
  title={Videotree: Adaptive tree-based video representation for llm reasoning on long videos},
  author={Wang, Ziyang and Yu, Shoubin and Stengel-Eskin, Elias and Yoon, Jaehong and Cheng, Feng and Bertasius, Gedas and Bansal, Mohit},
  booktitle={2025 IEEE/CVF Conference on Computer Vision and Pattern Recognition (CVPR)},
  pages={3272--3283},
  year={2025},
  organization={IEEE}
}

@inproceedings{atanasova2023faithfulness,
  title={Faithfulness tests for natural language explanations},
  author={Atanasova, Pepa and Camburu, Oana-Maria and Lioma, Christina and Lukasiewicz, Thomas and Simonsen, Jakob Grue and Augenstein, Isabelle},
  booktitle={Proceedings of the 61st Annual Meeting of the Association for Computational Linguistics (Volume 2: Short Papers)},
  pages={283--294},
  year={2023}
}

@inproceedings{whitehead2022reliable,
  title={Reliable visual question answering: Abstain rather than answer incorrectly},
  author={Whitehead, Spencer and Petryk, Suzanne and Shakib, Vedaad and Gonzalez, Joseph and Darrell, Trevor and Rohrbach, Anna and Rohrbach, Marcus},
  booktitle={European conference on computer vision},
  pages={148--166},
  year={2022},
  organization={Springer}
}

@inproceedings{yoon2025answerability,
  title={Can video llms refuse to answer? alignment for answerability in video large language models},
  author={Yoon, Eunseop and Yoon, Hee Suk and Hasegawa-Johnson, Mark and Yoo, Chang},
  booktitle={International Conference on Learning Representations},
  volume={2025},
  pages={86500--86525},
  year={2025}
}

@inproceedings{wang2024softsc,
  title={Soft self-consistency improves language models agents},
  author={Wang, Han and Prasad, Archiki and Stengel-Eskin, Elias and Bansal, Mohit},
  booktitle={Proceedings of the 62nd annual meeting of the association for computational linguistics (volume 2: Short papers)},
  pages={287--301},
  year={2024}
}

@article{lanham2023measuring,
  title={Measuring faithfulness in chain-of-thought reasoning},
  author={Lanham, Tamera and Chen, Anna and Radhakrishnan, Ansh and Steiner, Benoit and Denison, Carson and Hernandez, Danny and Li, Dustin and Durmus, Esin and Hubinger, Evan and Kernion, Jackson and others},
  journal={arXiv preprint arXiv:2307.13702},
  year={2023}
}

@inproceedings{fu2025videomme,
  title={Video-mme: The first-ever comprehensive evaluation benchmark of multi-modal llms in video analysis},
  author={Fu, Chaoyou and Dai, Yuhan and Luo, Yongdong and Li, Lei and Ren, Shuhuai and Zhang, Renrui and Wang, Zihan and Zhou, Chenyu and Shen, Yunhang and Zhang, Mengdan and others},
  booktitle={2025 IEEE/CVF Conference on Computer Vision and Pattern Recognition (CVPR)},
  pages={24108--24118},
  year={2025},
  organization={IEEE}
}

@article{factor2026,
  title={FACTOR: Counterfactual Training-Free Test-Time Adaptation for Open-Vocabulary Object Detection},
  author={Zhao, Kaixiang and Ye, Mao and Zhou, Lihua and Wang, Hu and Ji, Luping and Tang, Song and Zhu, Xiatian},
  journal={arXiv preprint arXiv:2605.03294},
  year={2026}
}

@article{kim2026ascmqra,
  title={Answer Self-Consistency with Margin-Triggered Question
         Re-Arbitration for the CVPR 2026 VidLLMs Challenge},
  author={Miyazawa, Tomoya and Okuno, Hiroyasu},
  journal={arXiv preprint arXiv:2606.04323},
  year={2026}
}

@article{rfeval2026,
  title={RFEval: Benchmarking reasoning faithfulness under counterfactual reasoning intervention in large reasoning models},
  author={Han, Yunseok and Lee, Yejoon and Do, Jaeyoung},
  journal={arXiv preprint arXiv:2602.17053},
  year={2026}
}

@article{agentsentry2026,
  title={Agentsentry: Mitigating indirect prompt injection in llm agents via temporal causal diagnostics and context purification},
  author={Zhang, Tian and Xu, Yiwei and Wang, Juan and Guo, Keyan and Xu, Xiaoyang and Xiao, Bowen and Guan, Quanlong and Fan, Jinlin and Liu, Jiawei and Liu, Zhiquan and others},
  journal={arXiv preprint arXiv:2602.22724},
  year={2026}
}

@article{mecd2024,
  title={MECD: Unlocking multi-event causal discovery in video reasoning},
  author={Chen, Tieyuan and Liu, Huabin and He, Tianyao and Chen, Yihang and Gan, Chaofan and Ma, Xiao and Zhong, Cheng and Zhang, Yang and Wang, Yingxue and Lin, Hui and others},
  journal={Advances in neural information processing systems},
  volume={37},
  pages={92554--92580},
  year={2024}
}


\appendix


\section{Proofs}
\label{app:proofs}

\paragraph{\Cref{thm:law}}
Condition on $(q,E,\hat a)$. Under \Cref{ass:matched}, the DESTROY
indicators
\[
X_j
=
\mathbb{1}
\left[
\hat a_{\mathrm{destroy}}^{(j)}
\neq
\hat a
\right]
\]
are i.i.d.\ $\operatorname{Bern}(\delta_d)$, and the SHAM indicators
\[
Y_j
=
\mathbb{1}
\left[
\hat a_{\mathrm{sham}}^{(j)}
\neq
\hat a
\right]
\]
are i.i.d.\ $\operatorname{Bern}(\delta_s)$. The two families are independent
because the conditions are executed as separate replays. Hence,
\[
X=\sum_{j=1}^{k}X_j
\sim
\operatorname{Bin}(k,\delta_d),
\qquad
Y=\sum_{j=1}^{k}Y_j
\sim
\operatorname{Bin}(k,\delta_s),
\]
independently, and
\[
\hat{\Delta}
=
\frac{X-Y}{k}
\]
takes values $m/k$ for integers $m\in\{-k,\ldots,k\}$.

Independence gives
\[
\Pr[X-Y=m]
=
\sum_j
\Pr[X=j+m]\Pr[Y=j],
\]
which is Eq.~\eqref{eq:law} in the notation
\[
b_i(p)
=
\binom{k}{i}p^i(1-p)^{k-i},
\]
with coefficients outside $\{0,\ldots,k\}$ treated as zero. Linearity gives
\[
\mathbb{E}
\left[
\hat{\Delta}
\right]
=
\delta_d-\delta_s
=
\Delta.
\]
Finally, independence and
\[
\operatorname{Var}
\left[
\frac{\operatorname{Bin}(k,p)}{k}
\right]
=
\frac{p(1-p)}{k}
\]
give
\[
\operatorname{Var}
\left[
\hat{\Delta}
\right]
=
\frac{
\delta_d(1-\delta_d)
+
\delta_s(1-\delta_s)
}{k},
\]
which is Eq.~\eqref{eq:variance}. Each term $p(1-p)$ is maximized at
$p=\tfrac12$, so
\[
\operatorname{sd}
\left[
\hat{\Delta}
\right]
\leq
\sqrt{\frac{1}{2k}},
\]
with equality when $\delta_s=\delta_d=\tfrac12$.

More generally, cross-branch dependence would introduce a covariance term:
\[
\operatorname{Var}
\left[
\hat{\Delta}
\right]
=
\operatorname{Var}
\left[
\hat{\delta}_d
\right]
+
\operatorname{Var}
\left[
\hat{\delta}_s
\right]
-
2\operatorname{Cov}
\left(
\hat{\delta}_d,
\hat{\delta}_s
\right).
\]
Positive covariance would reduce the variance relative to the independent
expression. Branch independence is therefore a property of the replay
implementation, not merely a mathematical convenience.
\hfill$\square$


\section{Additional primary results}
\label{app:additional-results}

\paragraph{Question-level reproducibility across replay seeds}
\Cref{tab:seed_question_level} reports pairwise agreement across the three
independent $k=3$ replay runs. Continuous scores are moderately correlated,
while strict trigger-set overlap is substantially lower.

\begin{table}[!t]
\centering
\caption{
\textbf{Question-level scores are moderately reproducible at $k=3$.}
All agreement statistics compare paired $\hat{\Delta}$ estimates for the
same LVBench questions across two independent replay runs. Pearson $r$ measures
continuous-score agreement; sign agreement measures whether paired scores fall
in the same negative, zero, or positive category; trigger-set Jaccard measures
overlap between the question sets selected by the strict
$\hat{\Delta}<0$ rule.
}
\small
\setlength{\tabcolsep}{3pt}
\renewcommand{\arraystretch}{1.15}

\begin{tabularx}{\columnwidth}{
    @{}
    >{\raggedright\arraybackslash}X
    >{\centering\arraybackslash}p{0.16\columnwidth}
    >{\centering\arraybackslash}p{0.20\columnwidth}
    >{\centering\arraybackslash}p{0.22\columnwidth}
    @{}
}
\toprule
Run pair &
Pearson $r$ &
Sign agreement &
Trigger-set Jaccard \\
\midrule
\runone\ vs.\ \runtwo
& 0.586 & 0.617 & 0.198 \\

\runone\ vs.\ \runthree
& 0.548 & 0.617 & 0.237 \\

\runtwo\ vs.\ \runthree
& 0.547 & 0.601 & 0.205 \\
\bottomrule
\end{tabularx}
\label{tab:seed_question_level}
\end{table}

\paragraph{Replay-budget diagnostics}
\Cref{tab:tie_budget} compares the observed sign distribution with the
homogeneous finite-replay prediction at $k=3$ and $k=10$, with Video-MME shown
as an external $k=3$ comparison.

\begin{table}[t]
\centering
\caption{
\textbf{Effect of replay budget on the distribution of estimated score
differences.} Proportions are computed over questions with a valid
$\hat{\Delta}$ estimate at the corresponding replay budget. For LVBench, the k=3 observed row corresponds to \runone, which is the run subsequently extended to $k=10$; Table 1 instead reports means across all three independent $k=3$ runs. Tie excess is the observed question-level tie rate minus the homogeneous finite-replay prediction.
}
\small
\setlength{\tabcolsep}{9pt}
\renewcommand{\arraystretch}{1.15}

\begin{threeparttable}
\begin{tabular}{clcccc}
\toprule
\kilth{Budget}
& \kilth{Distribution}
& {$\hat{\Delta}<0$}
& {$\hat{\Delta}=0$}
& {$\hat{\Delta}>0$}
& \kilth{Tie excess} \\
\midrule

\multicolumn{6}{l}{\textbf{LVBench}} \\
\addlinespace[2pt]

\multirow{2}{*}{$k=3$}
& Homogeneous prediction
& 0.117
& 0.236
& 0.647
& -- \\
& Observed (\runone)
& 0.133
& 0.298
& 0.569
& +0.062 \\

\addlinespace[4pt]

\multirow{2}{*}{$k=10$}
& Homogeneous prediction
& 0.056
& 0.071
& 0.874
& -- \\
& Observed 
& 0.165
& 0.123
& 0.712
& +0.052 \\

\addlinespace[3pt]
\cmidrule(lr){2-6}

& Change from $k=3$ to $k=10$
& +0.032
& $-$0.175
& +0.143
& $-$0.010 \\

\midrule

\multicolumn{6}{l}{\textbf{Video-MME}} \\
\addlinespace[2pt]

$k=3$
& Observed 
& 0.114
& 0.573
& 0.314
& -- \\

\bottomrule
\end{tabular}

\begin{tablenotes}[flushleft]
\footnotesize
\item
Tie excess is the observed tie probability minus the homogeneous-model
prediction. On LVBench, 67.2\% of questions tied at $k=3$ move to either side
of zero at $k=10$. Proportions are computed over questions with a valid
$\hat{\Delta}$ estimate at the corresponding budget. Video-MME provides
an external comparison at $k=3$.
\end{tablenotes}

\end{threeparttable}
\label{tab:tie_budget}

\end{table}

\paragraph{Component decomposition}
\Cref{tab:component-decomposition} separates the CARVE score into its
SHAM and DESTROY components across temporal-coverage and correctness strata.

\begin{table}[!t]
\centering
\caption{
\textbf{Component decomposition of the CARVE score by temporal coverage and
answer correctness.}
Entries are question-level mean SHAM and DESTROY flip rates within each
LVBench stratum; brackets give 95\% bootstrap confidence intervals obtained by
resampling videos. Among correct answers, the larger score for covered
questions is driven mainly by lower SHAM flip rates. Among incorrect answers,
the association reverses because covered questions have lower DESTROY flip
rates. The same qualitative pattern appears at both replay budgets.
}
\small
\setlength{\tabcolsep}{5pt}
\renewcommand{\arraystretch}{1.13}

\begin{tabular}{@{}llrcc@{}}
\toprule
\kilth{Replay budget} &
\kilth{Stratum} &
\kilth{$n$} &
\kilth{$\hat{\delta}_{s}$ (SHAM)} &
\kilth{$\hat{\delta}_{d}$ (DESTROY)} \\
\midrule

\multirow{4}{*}{$k=3$}
&
Covered, correct
&
275
&
\tablenum{0.215}\;{\footnotesize[0.181, 0.250]}
&
\tablenum{0.668}\;{\footnotesize[0.625, 0.710]}
\\

&
Uncovered, correct
&
293
&
\tablenum{0.281}\;{\footnotesize[0.241, 0.324]}
&
\tablenum{0.606}\;{\footnotesize[0.555, 0.658]}
\\

&
Covered, incorrect
&
245
&
\tablenum{0.420}\;{\footnotesize[0.376, 0.462]}
&
\tablenum{0.612}\;{\footnotesize[0.564, 0.657]}
\\

&
Uncovered, incorrect
&
436
&
\tablenum{0.424}\;{\footnotesize[0.391, 0.455]}
&
\tablenum{0.666}\;{\footnotesize[0.632, 0.700]}
\\

\midrule

\multirow{4}{*}{$k=10$}
&
Covered, correct
&
276
&
\tablenum{0.212}\;{\footnotesize[0.182, 0.242]}
&
\tablenum{0.644}\;{\footnotesize[0.605, 0.683]}
\\

&
Uncovered, correct
&
294
&
\tablenum{0.284}\;{\footnotesize[0.248, 0.322]}
&
\tablenum{0.617}\;{\footnotesize[0.574, 0.660]}
\\

&
Covered, incorrect
&
245
&
\tablenum{0.425}\;{\footnotesize[0.386, 0.463]}
&
\tablenum{0.593}\;{\footnotesize[0.545, 0.639]}
\\

&
Uncovered, incorrect
&
438
&
\tablenum{0.412}\;{\footnotesize[0.384, 0.438]}
&
\tablenum{0.665}\;{\footnotesize[0.634, 0.696]}
\\

\bottomrule
\end{tabular}
\label{tab:component-decomposition}
\end{table}


\section{Additional finite-replay analysis}
\label{app:finite-replay}

\paragraph{Discrete support at the primary replay budget.}
The support of $\hat{\Delta}$ is spaced in increments of $1/k$. At the
primary budget $k=3$, the score can take only seven values:
\[
-1,\;
-\frac{2}{3},\;
-\frac{1}{3},\;
0,\;
\frac{1}{3},\;
\frac{2}{3},\;
1.
\]
This coarse support creates a nonzero probability of exact empirical ties even
when the population score $\Delta$ is not zero. Increasing $k$ both reduces
the estimator variance through Eq.~\eqref{eq:variance} and creates a finer
score grid with spacing $1/k$. Consequently, questions tied at a low replay
budget can move to either side of zero when additional draws are observed.

\paragraph{Continuous agreement need not imply threshold agreement}
Thresholding discards information about score magnitude. For example,
\[
\hat{\Delta}^{(1)}
=
-\frac{1}{k},
\qquad
\hat{\Delta}^{(2)}
=
\frac{1}{k}
\]
differ by only $2/k$, but lie on opposite sides of the operational threshold.

More generally, suppose two replay runs produce
\[
\hat{\Delta}^{(2)}(q)
=
\hat{\Delta}^{(1)}(q)
+
\varepsilon(q),
\]
where $\varepsilon(q)$ is small relative to the total score range. The runs
may preserve much of the score ranking and therefore exhibit moderate or high
Pearson or Spearman correlation. Nevertheless, a question whose score lies
near zero can change binary status under a small perturbation. Selected-set
overlap can therefore be substantially lower than agreement between the
continuous scores.

The same reasoning applies when the replay budget changes. Scores at $k=3$
and $k=10$ can remain strongly correlated while the finer-budget estimate
moves low-budget ties or near-zero scores across the routing threshold.
Continuous-score stability and routing-set stability are therefore distinct
empirical properties.

\paragraph{Why resolving ties can weaken routing}
Consider the low-budget tied set
\[
S_0^{(k)}
=
\left\{
q:
\hat{\Delta}_k(q)=0
\right\}.
\]
When the budget increases, members of this set can resolve to either
$\hat{\Delta}>0$ or $\hat{\Delta}<0$. If the low-budget tied set
contains many questions that benefit from fallback reconsideration, then an
inclusive rule such as $\hat{\Delta}\leq0$ may perform well at low $k$.
Resolving those ties can shrink the selected set and reduce downstream gain,
even though the underlying score estimate has become more stable.

The finite-replay analysis therefore separates two questions:
\begin{enumerate}
    \item Does increasing $k$ improve estimation of the continuous score?
    \item Does the resulting score induce a better routing decision under the
    same threshold?
\end{enumerate}
The first follows from reduced variance and finer score-grid spacing. The
second is not determined by the estimator law and must be evaluated
empirically.

\paragraph{Scope of the analysis}
The analysis characterizes sampling behavior conditional on the
question-level replay probabilities $\delta_s$ and $\delta_d$. It does not
assume that all questions share one common pair of probabilities, and it does
not imply that finite-budget ties correspond to an exactly degenerate
population.

The analysis also does not establish that $\hat{\Delta}$ directly measures
temporal grounding or answer correctness. The score is a matched difference
between two answer-change rates. As \Cref{sec:e2} shows, its
relationship with annotated temporal coverage can arise through different
SHAM and DESTROY components across question strata.

Finally, lower estimator variance does not imply higher policy value under a
fixed threshold. \Cref{sec:e0} evaluates reproducibility across
independent replay seeds, \Cref{sec:e1} tests the effect of increasing
the replay budget from $k=3$ to $k=10$, and \Cref{sec:e3} evaluates
whether the selected questions benefit disproportionately from fallback
reconsideration.


\section{Secondary evaluations}
\label{app:secondary}

\subsection{Selective prediction on Video-MME}
\label{app:selective-prediction}

On the strict common-valid Video-MME subset ($n=595$), we rank retained
answers using $\hat{\Delta}$ and compare the resulting risk--coverage
curve with one based on sampled-position confidence. Because both methods
operate on the same evaluation set, they begin at 42.86\% accuracy under full
coverage.

At 90\% and 70\% coverage, ranking by $\hat{\Delta}$ retains a more
accurate subset. Over the 70--100\% coverage interval,
$\hat{\Delta}$ also attains the lower partial risk AUC, indicating better
selective performance when relatively few answers are withheld. Across the
full coverage range, however, confidence achieves the lower AURC because it
performs better under more aggressive abstention. CARVE therefore provides a
useful ordering near full coverage but does not uniformly dominate confidence.

\begin{table}[t]
\centering
\small
\setlength{\tabcolsep}{5pt}
\renewcommand{\arraystretch}{1.12}

\begin{tabular}{lcc}
\toprule
\kilth{Metric} &
$\hat{\Delta}$ &
\kilth{Confidence} \\
\midrule
Accuracy at 100\% coverage
& 42.86\% & 42.86\% \\

Accuracy at 90\% coverage
& 44.18\% & 43.47\% \\

Accuracy at 70\% coverage
& 44.81\% & 42.79\% \\

Partial risk AUC, 70--100\%
& 0.558 & 0.567 \\

Full-range AURC
& 0.544 & 0.521 \\
\bottomrule
\end{tabular}

\caption{
\textbf{Selective-prediction performance depends on the coverage regime.}
Results are computed over the strict common-valid Video-MME question subset
($n=595$). Coverage is the fraction of questions retained after ranking;
reported accuracies are computed over the retained questions at each coverage
level. Partial risk AUC summarizes the 70--100\% coverage range, while AURC
summarizes the full risk--coverage curve.
}
\label{tab:selective_prediction}
\end{table}

\subsection{Cross-backbone transfer}
\label{app:backbone-transfer}

We test whether a routing rule selected on one answerer remains useful when applied to another. The matched universe consists of the 837 LVBench questions (92 videos) with valid Phase-A trajectories, valid $k=1$ probe scores, and valid secondary-answerer (deferral) outputs under both the 7B and 32B backbone configurations. The experiment uses 400 video-level splits of this matched universe. For each split, the rule is selected using one source backbone and evaluated both on that backbone and after transfer to the other.

When selected on the 7B backbone, the rule yields a mean gain of
$+3.51\pm1.53$ percentage points on 7B and $+2.17\pm1.33$ points after
transfer to 32B. In the reverse direction, selection on 32B yields
$+1.95\pm1.21$ points on the source backbone and $+3.52\pm1.65$ points after
transfer to 7B. Mean transfer gain is therefore positive in both directions,
although its magnitude is asymmetric.

The inclusive $\hat{\Delta}\leq0$ policy is selected in 96.8\% of splits
with 7B as the source and 93.2\% with 32B as the source. The preferred
decision boundary is therefore more stable than the exact gain obtained after
transfer.

\begin{table}[t]
\centering
\small
\setlength{\tabcolsep}{3pt}
\renewcommand{\arraystretch}{1.15}

\begin{tabularx}{\columnwidth}{
    @{}
    >{\raggedright\arraybackslash}p{0.20\columnwidth}
    >{\centering\arraybackslash}p{0.23\columnwidth}
    >{\centering\arraybackslash}p{0.23\columnwidth}
    >{\centering\arraybackslash}X
    @{}
}
\toprule
\kilth{Source} &
\kilth{Source gain} &
\kilth{Transfer gain} &
\kilth{Rule selected} \\
&
(pp) &
(pp) &
(\%) \\
\midrule

7B
&
$+3.51\pm1.53$
&
$+2.17\pm1.33$
&
96.8
\\

32B
&
$+1.95\pm1.21$
&
$+3.52\pm1.65$
&
93.2
\\

\bottomrule
\end{tabularx}

\caption{
\textbf{The inclusive deferral rule transfers across answerer backbones.}
Entries report mean accuracy gain $\pm$ standard deviation over 400
independently sampled video-level splits. Source gain evaluates the selected
rule on the backbone used for rule selection; transfer gain evaluates that
same rule on the other backbone. Rule-selected percentages report the fraction
of the 400 splits in which the inclusive $\hat{\Delta}\leq0$ policy is
chosen.
}
\label{tab:backbone_transfer}
\end{table}


\section{A pre-flight procedure}
\label{sec:practice}

The finite-replay analysis and empirical diagnostics suggest a practical
procedure for evaluating a matched counterfactual probe before committing to
a full production run.

\paragraph{1. Measure replay variability under intact evidence}
Run SHAM on a pilot subset using a small replay budget and estimate
$\hat{\delta}_s$. A high SHAM flip rate indicates that rerendering and
re-answering intact evidence already produce substantial variability.
Equation~\eqref{eq:variance} can then be used to anticipate the scale of
per-question uncertainty at candidate values of $k$.

\paragraph{2. Verify that DESTROY adds an aggregate effect}
Run both SHAM and DESTROY and test whether the destroy--sham difference is
consistently positive across independent replay seeds. A large aggregate gap
shows that the destructive intervention changes answers beyond ordinary
replay variation, but does not establish stable question-level scores.

\paragraph{3. Inspect finite-budget discreteness and reproducibility}
Compare the observed negative, zero, and positive score fractions with the
finite-$k$ distribution implied by Eq.~\eqref{eq:law}. Exact ties are expected
at small $k$, but their prevalence and stability should be reported directly.
Repeat the pilot under multiple replay seeds and compare continuous scores,
score signs, and threshold-set overlap.

\paragraph{4. Test the intended replay budget directly}
When ties or threshold instability are operationally important, extend a
nested subset to a larger replay budget while preserving the original draws.
Measure both score stability and downstream policy behavior. Additional
replays should reduce estimator variance and refine the score grid, but need
not improve a fixed routing threshold.

\paragraph{5. Evaluate routing against matched controls}
When the score is used to trigger an intervention, compare the selected subset
against random subsets with the same cardinality and, where relevant, matched
video or evidence characteristics. Report repairs, harms, net yield, and the
selected policy's position within the matched-random distribution. This
separates the value of the fallback action from the value of the selection
rule.

Our experiments illustrate why each step matters. The aggregate
DESTROY--SHAM effect is highly reproducible, while question-level threshold
membership is less stable at $k=3$. Increasing the replay budget resolves most
low-budget ties and stabilizes the continuous score, yet weakens the original
$\hat{\Delta}\leq0$ routing policy. The matched-random analysis is
therefore necessary for determining whether a trigger set carries useful
routing information.


\section{Positioning among related methods}
\label{app:position}

\begin{table}[t]
\centering
\caption{
Comparison of CARVE with grounding, perturbation, and selective-prediction
methods. Train-free methods do not update the evaluated model; black-box
methods require no access to model internals; agentic methods operate over
multi-step tool-use trajectories.
}
\label{tab:novelty}

\fontsize{8.2}{9.6}\selectfont
\setlength{\tabcolsep}{3pt}
\renewcommand{\arraystretch}{1.12}

\begin{tabularx}{\columnwidth}{
@{}
>{\raggedright\arraybackslash}p{0.25\columnwidth}
>{\centering\arraybackslash}p{0.13\columnwidth}
>{\centering\arraybackslash}p{0.13\columnwidth}
>{\centering\arraybackslash}p{0.12\columnwidth}
>{\raggedright\arraybackslash}X
@{}
}
\toprule
\textbf{Method}
& \textbf{Train-free}
& \textbf{Black-box}
& \textbf{Agentic}
& \textbf{Primary signal} \\
\midrule

VER \citep{luo2025drift}
& \xmark
& \xmark
& \xmark
& Evidence-grounding reward \\

DeFacto \citep{xu2026defacto}
& \xmark
& \xmark
& \xmark
& Counterfactual supervision \\

VideoSEAL \citep{qiu2026videoseal}
& \xmark
& \xmark
& \cmark
& Temporal overlap and judge score \\

STEAR \citep{fan2026stear}
& \cmark
& \xmark
& \xmark
& Internal patch intervention \\

FACTOR \citep{factor2026}
& \cmark
& \cmark
& \xmark
& Attribute intervention \\

Self-consistency \citep{wang2023selfconsistency}
& \cmark
& \cmark
& \textit{Optional}
& Answer agreement \\

ASC-MQRA \citep{kim2026ascmqra}
& \cmark
& \cmark
& \cmark
& Agreement and re-watching \\

Reliable VQA \citep{whitehead2022reliable}
& \cmark
& \textit{Varies}
& \xmark
& Confidence and abstention \\

Video answerability \citep{yoon2025answerability}
& \xmark
& \xmark
& \xmark
& Learned answerability score \\

\midrule

\textbf{CARVE}
& \cmark
& \cmark
& \cmark
& \textbf{Intervention on retrieved evidence $E$} \\

\bottomrule
\end{tabularx}
\end{table}

Relative to VideoSEAL \citep{qiu2026videoseal}, CARVE requires neither gold
temporal intervals nor an external judge to compute its score, and it audits a
frozen agent without architectural modification or retraining. Relative to
STEAR \citep{fan2026stear}, CARVE intervenes on externally observable evidence
selected by the agent rather than on internal layers or patch representations.
Relative to DeFacto \citep{xu2026defacto}, CARVE is a test-time measurement
for video agents rather than a training objective for image-based reasoning.
Relative to FACTOR \citep{factor2026}, CARVE intervenes on evidence gathered
through an agent trajectory rather than on attributes of a single test image.

CARVE is also distinct from agreement-, confidence-, and judge-based
selectors. Those methods estimate reliability from repeated answers, output
uncertainty, or an evaluator's assessment. CARVE instead measures how the
committed answer responds when the agent's retrieved visual evidence is
destroyed relative to a matched intact-evidence replay.


\section{Reproducibility details}
\label{app:repro}

\paragraph{Stored artifacts}
For each eligible question, we retain the frozen trajectory, parsed original
answer, reconstructed evidence locations and timestamps, rendering prompt,
SHAM and DESTROY replay outputs, parsed replay answers, validity indicators,
flip indicators, replay seeds, and phase-randomized realizations.

\paragraph{Fixed configuration}
Prompts, decoding settings, parsing rules, context limits, retrieval
parameters, tool parameters, and policy thresholds are fixed within each
production experiment.

\paragraph{Replay experiments}
The primary LVBench probe is replicated across three independent $k=3$ runs.
The replay-budget experiment extends the primary run from $k=3$ to $k=10$
while preserving the original three draws and adding seven new draws per
condition.

\paragraph{Statistical analysis}
LVBench confidence intervals and hypothesis tests are clustered by video.
The cross-backbone experiment uses 400 video-level splits with identical split
seeds across comparisons and selects the routing rule using source development
videos only.

\paragraph{Software and hardware}
Production runs use NVIDIA RTX 4090 GPUs. The serving environment uses
vLLM~0.8.4, PyTorch~2.6.0 with CUDA~12.4, and
Transformers~4.51.3.

\FloatBarrier

\end{document}